\documentclass[times, review, 10pt]{elsarticle}

\usepackage{amssymb}
\usepackage{amsmath}
\usepackage{url}
\usepackage{booktabs}
\usepackage{float}  
\usepackage{graphicx}
\usepackage{subcaption}
\usepackage{multirow} 
\usepackage{makecell}
\usepackage{array}

\journal{Pattern Recognition}

\begin{document}

\begin{sloppypar}

\begin{frontmatter}



\title{InceptionMamba: An Efficient Hybrid Network with Large Band Convolution and Bottleneck Mamba}

\renewcommand{\thefootnote}{*}
\author{Yuhang Wang$^{1}$, Jun Li$^{1}$\footnote{Corresponding author (lijuncst@njnu.edu.cn).}, Zhijian Wu$^{2}$, Jifeng Shen$^{3}$, Jianhua Xu$^{1}$, Wankou Yang$^{4}$} 

\address{$^{1}$ School of Computer and Electronic Information, Nanjing Normal University, Nanjing, China\\
$^{2}$ Medical Artificial Intelligence Lab, Westlake University, Zhejiang, China \\
$^{3}$ School of Electrical and Information Engineering, Jiangsu University, Zhenjiang, China \\
$^{4}$ School of Automation, Southeast University, Nanjing, China \\
}

\begin{abstract}

Within the family of convolutional neural networks, InceptionNeXt has shown excellent competitiveness in image classification and a number of downstream tasks. Built on parallel one-dimensional strip convolutions, however, it suffers from limited ability of capturing spatial dependencies along different dimensions and fails to fully explore spatial modeling in local neighborhood. Additionally, although InceptionNeXt excels at multi-scale spatial modeling capacity, convolution operations are prone to locality constraints and inherently lack the global receptive field, which is detrimental to effective global context modeling. Recent research demonstrates that Mamba based on state-space models can capture long-range dependencies while enjoying linear computational complexity. Thus, it is necessary to take advantage of the Mamba architecture to improve the long-range modeling capabilities of the InceptionNeXt while maintaining desirable efficiency. Towards this end, we propose a novel backbone architecture termed InceptionMamba in this study. More specifically, the traditional one-dimensional strip convolutions are replaced by orthogonal band convolutions in our InceptionMamba to achieve cohesive spatial modeling. Furthermore, global contextual modeling can be achieved via a bottleneck Mamba module, facilitating enhanced cross-channel information fusion and enlarged receptive field. Extensive evaluations on classification and various downstream tasks demonstrate that the proposed InceptionMamba achieves state-of-the-art performance with superior parameter and computational efficiency. The source code will be available at \url{https://github.com/Wake1021/InceptionMamba}.
\end{abstract}


\begin{keyword}
InceptionMamba \sep Orthogonal Band Convolutions \sep Global Contextual Modeling \sep Bottleneck Mamba \sep Cross-channel Information Fusion 



\end{keyword}

\end{frontmatter}


\begin{figure}[htbp]
    \centering
    \begin{subfigure}{0.49\textwidth}
        \centering
        \includegraphics[width=\linewidth]{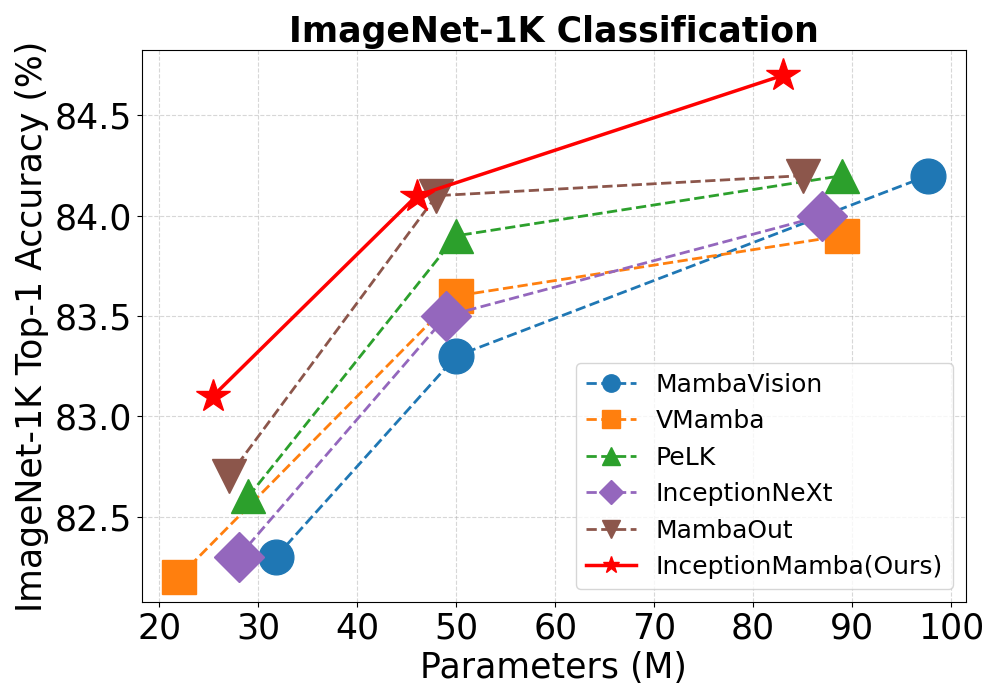}
        \caption{Params. vs Top-1(\%)}
        \label{fig:sub1}
    \end{subfigure}
    \hfill 
\begin{subfigure}{0.49\textwidth}
        \centering
        \includegraphics[width=\linewidth]{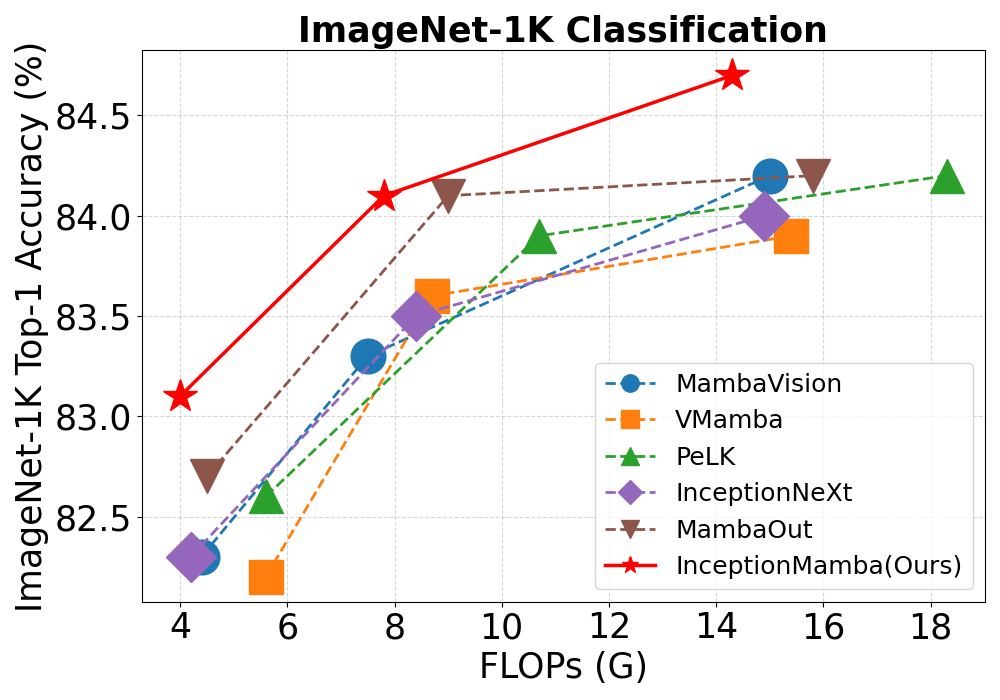}
        \caption{FLOPs vs Top-1(\%)}
        \label{fig:sub2}
    \end{subfigure}
    \caption{Performance on ImageNet-1K validation set at $224^2$ resolutions.}
    \label{Fig:1}
\end{figure}

\section{Introduction}

In Convolutional Neural Networks (CNNs)~\cite{lecun1989backpropagation, lecun1998gradient, liu2022convnet, He2015,DenseNet2017}, the success of Inception architecture is largely attributed to efficient strip convolution, evolving from local feature extraction to dynamic parameter optimization~\cite{liu2022convnet}. Originated from InceptionNeXt~\cite{yu2024inceptionnext}, it is beneficial for reducing the number of parameters through asymmetric convolution decomposition (such as splitting a 7×7 convolution into 1×7 and 7×1 strip convolutions). As a representative member within Inception family, InceptionNeXt has showcased the efficacy of strip convolutions in balancing computational efficiency and representational abilities. In InceptionNeXt, large-kernel depthwise convolutions are decomposed into four efficient branches including parallel one-dimensional strip convolutions, such that large-kernel CNNs are accelerated with maintained performance.

Despite achieving huge success, conventional one-dimensional strip convolutions in InceptionNeXt suffer twofold inherent limitations, which hinder their potential in complex visual understanding scenarios. On the one hand, the one-dimensional strip convolutions are independently computed in separate channel groups, lacking sufficient spatial modeling capacity in capturing complex visual contents. For instance, while horizontal kernels effectively characterize axis-aligned patterns, their weakened responses to orthogonal structures (e.g., vertical edges) results in less discriminative encoding. Additionally, the local inductive bias of CNNs, arising from utilizing limited-size convolutional kernels (e.g., 3$\times$3 convolutions) for hierarchical local feature extraction, fundamentally constrains their capacity in global contextual relationship modeling, thereby suffering from restricted receptive field. While recent efforts to addressing this limitation have explored oversized kernels~\cite{ding2022scalingkernels31x31revisiting} and dilated convolutions~\cite{yu2016multi} to expand receptive fields, these approaches are challenged by accuracy-efficiency trade-off: enlarged kernels introduce a quadratic increase in computational complexity $(O(k^2))$, while dilated convolutions are prone to grid artifacts that degrade spatial coherence in feature representations. In a nutshell, CNNs like InceptionNeXt with local receptive fields and weight-sharing mechanisms are highly efficient at capturing local cues but still struggle to model long-range dependencies. Consequently, it is necessary to improve the long-range modeling capacity of the InceptionNeXt while maintaining efficient parallel structure for enlarged receptive field.

Recently, a new architecture, i.e., Vision Transformer (ViT)~\cite{dosovitskiy2020vit,tolstikhin2021mixer,steiner2021augreg,chen2021outperform,zhuang2022gsam,zhai2022lit}, has received extensive attention and emerged as a promising alternative to CNNs. ViT leverages self-attention mechanisms to enable global interactions, yet it incurs quadratic computational complexity, which poses enormous challenges to computationally intensive scenarios, e.g., dense prediction tasks. Similar to ViT, the Mamba architecture based on State Space Models has appealing long-range modeling capacity superior to CNNs. It employs a selective state transition mechanism that combines the linear complexity of convolutions with data-dependent global context modeling. Additionally, the gating mechanism allows the Mamba to dynamically focus on key spatial positions while preserving sub-quadratic computational complexity. This attribute well addresses the spatial modeling limitations of the InceptionNeXt, while hardly incurring expensive computational costs like ViT.


\begin{figure*}
    \centering
    \includegraphics[width=1.0\textwidth]{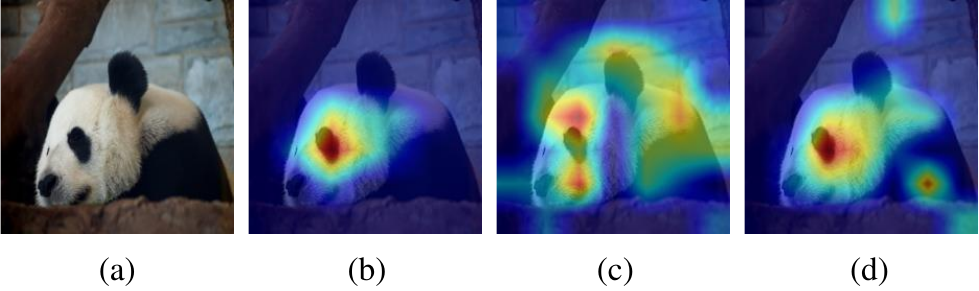}
    \caption{Visualizations of different CAMs~\cite{selvaraju2017grad} generated from ConvNeXt(b), InceptionNeXt(c) and our InceptionMamba(d) given a panda image(a). It can be observed that InceptionMamba showcases superior spatial modeling capacity in capturing semantic-aware areas in the neighborhood of panda eye and body. In contrast, InceptionNeXt only focuses on scattered eye-centric regions, yielding less cohesive representations.}
    \label{Fig:2}
\end{figure*}

Taking advantage of global contextual modeling capability of Mamba, we propose an efficient hybrid backbone architecture with large band convolutions and bottleneck Mamba termed InceptionMamba in this study. Inherited from Inception architecture, our InceptionMamba maintains Inception-style depthwise convolutions with efficient parallel structures, yet features carefully designed orthogonal band convolutions instead of the traditional one-dimensional stripe convolutions. Thus, multi-dimensional visual cues can be discovered for cohesive spatial modeling, leading to comprehensive embedding with improved representational capacity. In addition, to improve inter-channel interaction ability, we incorporate a bottleneck Mamba module into the token mixture for facilitating cross-channel information integration and enlarged receptive field. The advantages of our framework are illustrated in Fig.~\ref{Fig:2}, demonstrating that InceptionMamba can capture semantically correlated local neighborhood with sufficient discriminative power. For instance, our method can accurately focus on object-aware regions surrounding panda eye and body, whereas InceptionNeXt only identify isolated eye-centric areas with degraded spatial modeling capacity. 

The superiority of our InceptionMamba has been verified through extensive evaluations on various public benchmarking datasets for both classification and other downstream tasks. As shown in Fig.~\ref{Fig:1}, our InceptionMamba-B attains 84.7\% top-1 accuracy on ImageNet-1K, outperforming state-of-the-arts including both InceptionNeXt and Mamba-like frameworks (e.g., MambaVision~\cite{hatamizadeh2025mambavision} and VMamba~\cite{liu2024vmamba}), while enjoying competitive computational efficiency compared to InceptionNeXt.

To summarize, our contributions in this study are fourfold as follows:
\begin{itemize}
    \item We propose a hybrid backbone architecture termed InceptionMamba with efficient parallel structure inherited from Inception and develop a model family accordingly.
    \item In contrast to one-dimensional strip convolution involved in InceptionNeXt, our carefully designed orthogonal band convolutions significantly contribute to characterizing semantically cohesive visual cues in local adjacent neighborhood, thereby demonstrating improved spatial modeling capacity. 
    \item The devised bottleneck Mamba module is conducive to inter-channel information fusion, while expanding the receptive field for enhancing global contextual modeling.
    \item Extensive evaluations on public benchmarking datasets for classification and other downstream tasks demonstrate that our InceptionMamba achieves state-of-the-art performance with competitive efficiency.
    
\end{itemize}

\section{Related Work}

\subsection{Inception Architecture}


The evolution of the Inception architecture began with the core idea of multi-scale feature fusion. Earliest Inception architecture~\cite{szegedy2015going,szegedy2016CVPR} introduces parallel multi-branch convolutions (1$\times$1, 3$\times$3, 5$\times$5, and pooling) along with 1$\times$1 dimensionality reduction to overcome the single-scale limitation. Subsequent improvements of Inception-v2~\cite{szegedy2016rethinking} and Inception-v3~\cite{szegedy2016rethinking} are manifested in enhancing computational efficiency via convolution decomposition (e.g., splitting one 5$\times$5 convolution into two 3×3 convolutions) and batch normalization (BN). Similar pure Inception variant is Inception-v4~\cite{szegedy2017inception} which contains multiple consecutive Inception blocks with parallel structures. More advanced Inception-ResNet~\cite{ding2021diversebranchblockbuilding, yu2024inceptionnext} is a hybrid Inception network which integrates the residual learning into Inception architecture, achieving network training speedup through skip connections. A milestone work within Inception family emerges as InceptionNeXt~\cite{yu2024inceptionnext} which decomposes large-kernel depthwise convolutions into multiple parallel branches of smaller kernels (e.g., square 3$\times$3 convolutions and one-dimensional strip convolutions), achieving desirable accuracy-throughputs trade-off in a series of downstream tasks. It is significantly beneficial for accelerating large-kernel networks with maintained performance. 

\subsection{Mamba}

As a sequence modeling framework inspired by control theory, Mamba~\cite{mamba, mamba2} is a new State Space Model (SSM) which is capable of capturing long-range dependencies through the evolution of hidden states. The continuous system is defined by three core matrices: the state transition matrix $ \mathbf{A} \in \mathbb{R}^{M \times M} $, the input projection matrix \( \mathbf{B} \in \mathbb{R}^{M \times 1} \), and the output projection matrix \( \mathbf{C} \in \mathbb{R}^{1 \times M} \). The system dynamics can be described as follows:
\begin{equation}
    \begin{aligned}
        h^{'}(t) &= \mathbf{A}h(t)+\mathbf{B}x(t) \\
        y(t) &= \mathbf{C}h(t) \\
    \end{aligned}
\end{equation}
To achieve efficient discrete computation, Mamba introduces zero-order preserving discretization method, which converts the continuous system into a recursive form through the time scale parameter $\Delta$ :
\begin{equation}
    \begin{aligned}
        \mathbf{\overline{A}} &= e^{\Delta \mathbf A} \\
        \quad \mathbf{\overline{B}} &= ({\Delta \mathbf A})^{-1}(e^{{\Delta \mathbf A}} - \mathbf{I}) \cdot {\Delta \mathbf B}
    \end{aligned}
\end{equation}
Thus, the discrete state update equation is obtained:
\begin{equation}
    \begin{aligned}
            h_t &= \mathbf{\overline{A}} h_{t-1} + \mathbf{\overline{B}} x_t \\
        \quad y_t &= \mathbf{C} h_t 
    \end{aligned}
\end{equation}
Retaining the sequence modeling characteristics of RNN, the above formulation can be reconstructed in a parallel computing manner from the global convolution perspective. The system output can be equivalently expressed as:
\begin{equation}
    \begin{aligned}
         \mathbf{\overline{K}} &= (\mathbf{C\overline{B}},\mathbf{C\overline{AB}}, \dots, \mathbf{C\overline{A}^{L-1}\overline{B}}) \\
          \mathbf{y} &= \mathbf{x} \ast \overline{\mathbf{K}} 
    \end{aligned}
\end{equation}
where $\ast$ represents the convolution operation, and $\overline{\mathbf{K}} \in \mathbb{R}^{L}$  serves as the kernel of the SSM.

Recent advancements in visual Mamba models demonstrate the potential of SSM in a variety of vision tasks. In Vision Mamba~\cite{vim}, a bidirectional Mamba block was proposed for efficient high-resolution image processing. VMamba introduces Cross-Scan strategy to strengthen global context modeling. MambaVision combines linear complexity of Mamba with Transformer self-attention in a hybrid architecture to enhance efficiency and global spatial reasoning. Recent research explores the essence of Mamba, and reveals that Mamba is not ideally suited for all vision tasks. Meanwhile, a range of MambaOut~\cite{yu2025mambaout} models are obtained by stacking Mamba blocks while removing their core token mixer SSM. 

\begin{figure*}
	\centering
       \includegraphics[width=1.0\textwidth]{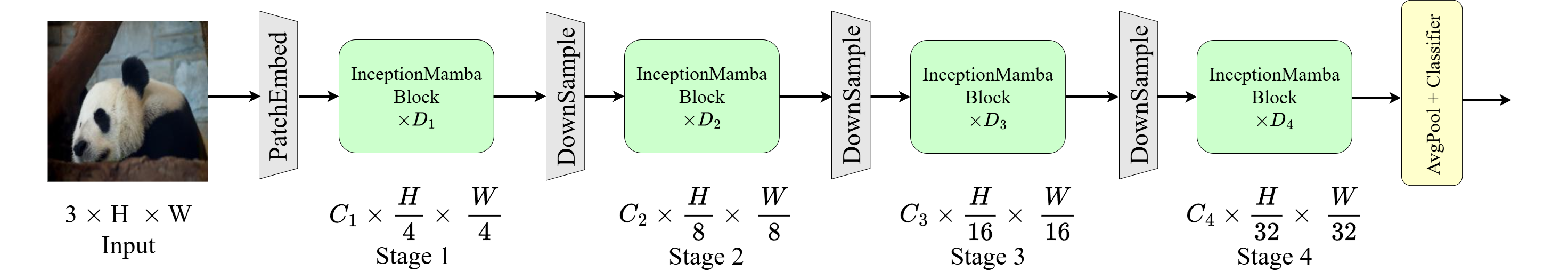}  
	\caption{\textbf{Our InceptionMamba architecture with four stages.} Similar to ConvNeXt~\cite{liu2022convnet}, InceptionMamba employs hierarchical architecture of four consecutive stages. Each stage consists of a patch embedding layer or a downsampling module, combined with $N_i$ InceptionMamba blocks.}  
	\label{Fig:3}
\end{figure*}

\section{Method}
Encouraged by recent success of vision Mamba models, we propose to deeply integrate Mamba SSM into efficient InceptionNeXt architecture with parallel structures. The former, which enjoys linear-complexity sequence modeling capability, can capture long-range spatial dependencies, whilst the latter excels at extracting key features while maintaining network efficiency. In this sense, our synergistic design helps to preserve strengths of CNNs in efficient local reasoning while enhancing model adaptability to complex scenes through global dynamic feature selection mechanism of Mamba, establishing a new foundation paradigm for a series of downstream vision tasks.

\subsection{Overview of InceptionMamba}

Inspired by InceptionNeXt, we have designed a four-stage InceptionMamba architecture as illustrated in Fig.~\ref{Fig:3}. For a given input tensor $X\in \mathbb{R}^{B \times C \times H \times W}$ where $B$, $C$, $H$, $W$ represent the batch size, number of channels, height, and width, respectively, the feature processing begins with a ConvMixer module that emphasizes local spatial information encoding similar to token mixer in InceptionNeXt:
\begin{equation}
\begin{aligned}
X^{'} = \text{ConvMixer}(X)
\end{aligned}
\end{equation}

The output $X^{'}$ is then forwarded to a Global Mixer module for global context modeling and capturing long-range dependencies:
\begin{equation}
\begin{aligned}
X^{''} = \text{GlobalMixer}(X^{'})
\end{aligned}
\end{equation}

Then, normalization is applied to the aggregated feature expressed as:
\begin{equation}
\begin{aligned}
Y = \text{Norm}(X^{''})
\end{aligned}
\end{equation}
After normalization, the resulting feature $Y$ is passed through a MLP module consisting of two fully connected layers, with a non-linear activation function inserted in between. These fully connected layers are typically implemented as $ 1 \times 1 $ convolutions for efficiency. Additionally, a residual connection is employed to facilitate gradient flow and model stability. This process is formulated as:
\begin{equation}
\begin{aligned}
Y &= \text{Conv}_{1\times1}^{rC \to C}\{\sigma[\text{Conv}_{1\times1}^{C \to rC}(Y)]\} + X  \\
\end{aligned}
\end{equation}
where $\text{Conv}_{k\times k}^{C_{i} \to C_{o}}$ denotes a convolution operation with kernel size $k \times k $. Input and output channels are denoted as $C_i$ and $C_o$, respectively. Additionally, $r$ is the expansion ratio and $\sigma$ denotes a non-linear activation function.


As the cornerstone of our method, the core InceptionMamba block constitute four-stage framework resembling ConvNeXt and InceptionNeXt, leading to a family of InceptionMamba model with varying sizes. More specifically, the block numbers in the three model variants are [3, 3, 12, 3], [4, 4, 32, 4] and [4, 4, 34, 4] for the Tiny (T), Small (S) and Base (B), respectively. Starting from an input image of $3 \times H \times W$, our model first applies patch embedding to process the visual data. In each stage, subsequently, downsampling operation is combined with multiple InceptionMamba blocks to progressively reduce spatial resolution, i.e., spatial size $H \times W$ is reduced by 4 times, 16 times, 32 times, and 64 times respectively, while expanding the channel dimension (C). This pyramid-like hierarchical architecture ends with global average pooling and classifiers for the final prediction. The architecture configurations of the three InceptionMamba variants can be referred to Table~\ref{Tab:1}.

\begin{table}[htbp]
\centering
\caption{Architectures of our InceptionMamba family including Tiny (T), Small (S) and Base (B) versions. }  
\small
\renewcommand{\arraystretch}{1.1} 
\setlength{\tabcolsep}{2pt} 
\begin{tabular}{
    c   |  
    c   |  
    c   |   
    c   |   
    >{\centering\arraybackslash}p{1.5cm} | 
    >{\centering\arraybackslash}p{1.5cm} |  
    >{\centering\arraybackslash}p{1.5cm}   
}
\hline
\multirow{2}{*}{Stage} & \multirow{2}{*}{\#Tokens} & \multicolumn{2}{c|}{\multirow{2}{*}{Layer Specification}} & \multicolumn{3}{c}{InceptionMamba} \\
\cline{5-7}
 &  & \multicolumn{2}{c|}{} & T & S & B \\
\hline
\multirow{6}{*}{1} & \multirow{6}{*}{$\frac{H}{4} \times \frac{W}{4}$} & \multirow{2}{*}{PatchEmbed} & Kernel Size & \multicolumn{3}{c}{$3 \times 3$, stride 2} \\
\cline{4-7}
 &  &  & Embed. Dim. & \multicolumn{2}{c|}{72} & 96 \\
\cline{3-7}
 &  & \multirow{5}{*}{\makecell{InceptionMamba \\ Block}} & Kernel size & \multicolumn{3}{c}{$3 \times 3, 3 \times 11, 11 \times 3$} \\
\cline{4-7}
 &  &  & Conv. group ratio & \multicolumn{3}{c}{1/8} \\
\cline{4-7}
 &  &  & Bottleneck ratio & \multicolumn{3}{c}{1/2} \\
\cline{4-7}
 &  &  & MLP Ratio & \multicolumn{3}{c}{4} \\
\cline{4-7}
 &  &  & \# Block & 3 & \multicolumn{2}{c}{4} \\
\hline
\multirow{6}{*}{2} & \multirow{6}{*}{$\frac{H}{8} \times \frac{W}{8}$} & \multirow{2}{*}{DownSampling} & Kernel Size & \multicolumn{3}{c}{$3 \times 3$, stride 2} \\
\cline{4-7}
 &  &  & Embed. Dim. & \multicolumn{2}{c|}{144} & 192 \\
\cline{3-7}
 &  & \multirow{5}{*}{\makecell{InceptionMamba \\ Block}} & Kernel size & \multicolumn{3}{c}{$3 \times 3, 3 \times 11, 11 \times 3$} \\
\cline{4-7}
 &  &  & Conv. group ratio & \multicolumn{3}{c}{1/8} \\
\cline{4-7}
 &  &  & Bottleneck ratio & \multicolumn{3}{c}{1/2} \\
\cline{4-7}
 &  &  & MLP Ratio & \multicolumn{3}{c}{4} \\
\cline{4-7}
 &  &  & \# Block & 3 & \multicolumn{2}{c}{4} \\
\hline
\multirow{6}{*}{3} & \multirow{6}{*}{$\frac{H}{16} \times \frac{W}{16}$} & \multirow{2}{*}{DownSampling} & Kernel Size & \multicolumn{3}{c}{$3 \times 3$, stride 2} \\
\cline{4-7}
 &  &  & Embed. Dim. & \multicolumn{2}{c|}{288} & 384 \\
\cline{3-7}
 &  & \multirow{5}{*}{\makecell{InceptionMamba \\ Block}} & Kernel size & \multicolumn{3}{c}{$3 \times 3,3 \times 11, 11 \times 3$} \\
\cline{4-7}
 &  &  & Conv. group ratio & \multicolumn{3}{c}{1/8} \\
\cline{4-7}
 &  &  & Bottleneck ratio & \multicolumn{3}{c}{1/2} \\
\cline{4-7}
 &  &  & MLP Ratio & \multicolumn{3}{c}{4} \\
\cline{4-7}
 &  &  & \# Block & 12 & 32 & 34 \\
\hline
\multirow{6}{*}{4} & \multirow{6}{*}{$\frac{H}{32} \times \frac{W}{32}$} & \multirow{2}{*}{DownSampling} & Kernel Size & \multicolumn{3}{c}{$3 \times 3$, stride 2} \\
\cline{4-7}
 &  &  & Embed. Dim. & \multicolumn{2}{c|}{576} & 768 \\
\cline{3-7}
 &  & \multirow{5}{*}{\makecell{InceptionMamba \\ Block}} & Kernel size & \multicolumn{3}{c}{$3 \times 3, 3 \times 11, 11 \times 3$} \\
\cline{4-7}
 &  &  & Conv. group ratio & \multicolumn{3}{c}{1/8} \\
\cline{4-7}
 &  &  & Bottleneck ratio & \multicolumn{3}{c}{1/2} \\
\cline{4-7}
 &  &  & MLP Ratio & \multicolumn{3}{c}{4} \\
\cline{4-7}
 &  &  & \# Block & 3 & \multicolumn{2}{c}{4} \\
\hline
 \multicolumn{7}{c}{Global average pooling, MLP} \\
\hline
\multicolumn{4}{c|}{Parameters (M)} & 25 & 46 & 83 \\
\hline
\multicolumn{4}{c|}{FLOPs (G)} & 4.0 & 7.8 & 14.3 \\
\hline
\end{tabular}
\label{Tab:1} 
\end{table}

\begin{figure}[t]
	\centering
    \includegraphics[scale=0.2]{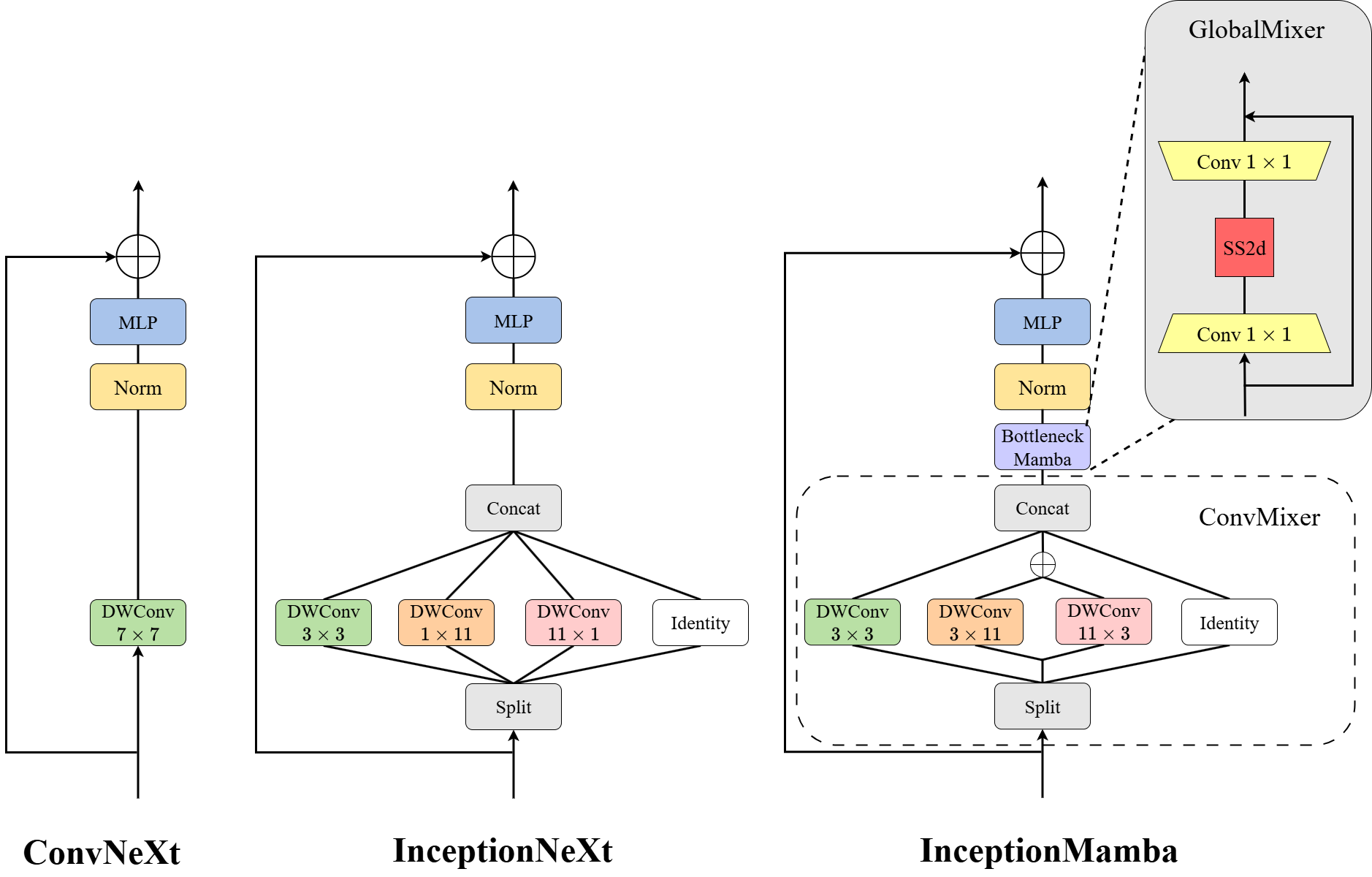}  
	\caption{Illustrative comparison of ConvNeXt, InceptionNeXt, and our InceptionMamba block. InceptionMamba consists of two key components, namely ConvMixer and GlobalMixer. The former inherits efficient parallel structure from InceptionNeXt to accelerate large-kernel depthwise convolutions. Different from InceptionNeXt using one-dimensional strip convolutions, our InceptionMamba takes advantage of orthogonal band convolutions to capture multi-dimensional visual cue for cohesive feature embedding. Furthermore, the GlobalMixer adopts a carefully designed bottleneck Mamba to facilitate inter-channel information with enlarged receptive field. Compared to ConvNeXt and InceptionNeXt, our InceptionMamba enjoys efficient local spatial modeling capacity while excels in capturing long-range dependencies.}  
	\label{Fig:4}
\end{figure}

\subsection{ConvMixer with large-kernel band convolutions}

Inspired by SLaK~\cite{liu2022more} and InceptionNeXt, we devise a multi-branch structure in our ConvMixer to address the limitation of one-way large kernels. Inherited from InceptionNeXt block, the parallel structure allows speedup of large-kernel depthwise convolutions for efficient local spatial modeling. Different from its predecessor, however, our ConvMixer employs orthogonal large-kernel band convolutions instead of one-dimensional strip convolutions in InceptionNeXt. Band convolution can focus on larger areas with improved local modeling capability. This advantage can be intuitively shown in Fig.~\ref{Fig:2}, clearly demonstrating that our InceptionMamba generates more spatially cohesive activations on the object-aware regions compared to InceptionNeXt.

Mathematically, the input features $X$ are divided into three groups along the channel dimension via our ConvMixer as follows:
\begin{equation}
    X_{square}, X_{band}, X_{identity} = Split(X)
\end{equation}
which are subsequently fed into three different branches for parallel processing:
\begin{equation}\label{equ10}
    \begin{aligned}
        X_{1} &= DWConv_{3 \times 3}(X_{square}) \\
        X_{2} &= DWConv_{3 \times 11}(X_{band}) + DWConv_{11 \times 3}(X_{band}) \\
        X_{3} &= X_{id} \\
    \end{aligned}
\end{equation}

In Eq.~(\ref{equ10}), the sizes of orthogonal large-kernel band convolutions are $3\times11$ and $11\times3$. Finally, the output features of each branch are concatenated as:
\begin{equation}
    X^{'}=Concat(X_1, X_2, X_3)
\end{equation}

Notably, identity mapping operations are applied to the majority of channel groups in the ConvMixer while depthwise convolutions are only imposed on a small subset of groups. This is consist with InceptionNeXt which reveals that depthwise convolutions are not required for most channels. 

\subsection{GlobalMixer with bottleneck Mamba}

\begin{figure}
    \centering
    \includegraphics[width=1.0\textwidth]{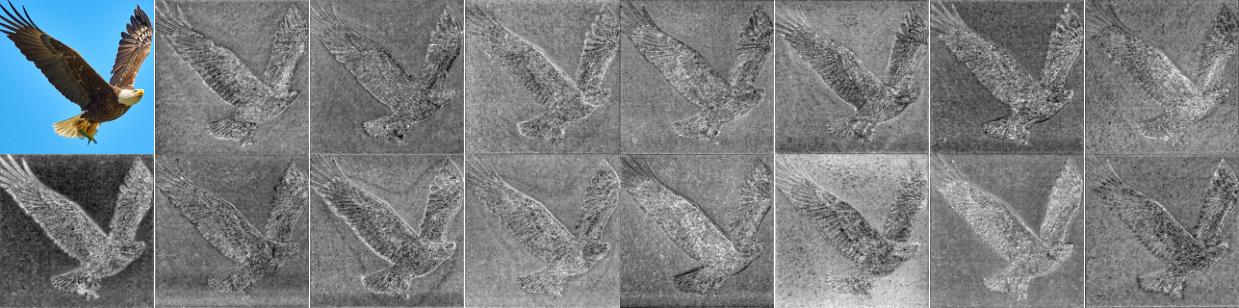}
    \caption{Visualized feature maps along consecutive channels in an intermediate layer of a pretrained VMamba model. The top-left image shows the input. Substantial redundancies in channel information can be clearly observed.}
    \label{Fig:5}
\end{figure}

In the traditional InceptionNeXt architecture, the application of depthwise separable convolutions and identity mappings leads to limited channel interaction without sufficiently exploring long-range dependency. Moreover, existing deep architectures commonly suffer from redundant channel information as illustrated in Fig.~\ref{Fig:5}, which inevitably increases the model complexity. To address this limitation, we design a Bottleneck Mamba structure to enhance effective cross-channel interaction and reduce channel redundancy, thereby enhancing the model capacity while hardly affecting the model efficiency. Towards this end, we design a GlobalMixer module for inter-channel interaction at the Token layer after the ConvMixer module. Traditional methods including ConvNeXt and InceptionNeXt only pass the concatenated features through a straightforward MLP, leading to inadequate cross-channel interactions due to inherent single-channel processing characteristic of depthwise convolution. To mitigates this drawback, a bottleneck structure is introduced into the GlobalMixer, making use of $1 \times 1$ convolution to compress and expand feature channels. This is beneficial for efficient cross-channel fusion in a low-dimensional space while retaining the key information of the original features.

To strengthen global modeling capacities while optimizing computational efficiency, a state-space module (SS2D)is integrated within the bottleneck architecture. This co-design leverages channel compression mechanism (ratio $r=2$) of the bottleneck for significantly reducing the SS2D dimensionality. More importantly, the SS2D module can capture long-range dependencies with the help cross-scan mechanism, while the compressed feature space enjoys desirable parameter efficiency. Mathematically, given the Input $X^{'}$, our GlobalMixer with a bottleneck Mamba is formulated as follows:

\begin{equation}
\begin{aligned}
X^{''} &= \text{Conv}_{1\times1}^{C \to C/r}(X^{'}))  \\
X^{''} &= \text{SS2D}(X^{''})  \\
X^{''} &= \text{Conv}_{1\times1}^{C/r \to C}(X^{''}) \\
Y &= X + X^{'}
\end{aligned}
\label{eq:bottleneck}
\end{equation}

Benefiting from our efficient GlobalMixer, InceptionMamba achieves a 13.8\% parameter reduction (29M$\rightarrow$25M) and 15.2\% FLOPs reduction (4.6G$\rightarrow$4.0G) while preserving 83.1\% ImageNet top-1 accuracy compared to directly utilizing SS2D for channel fusion, suggesting its potential in serving as a promising paradigm for balancing representational capacity with hardware efficiency.

\subsection{Block Comparisons}
Since our InceptionMamba is closely related to ConvNeXt and InceptionNeXt frameworks, we compare different building blocks comprising the three frameworks as illustrated in Fig.~\ref{Fig:4}. With the help of the ConvMixer, our InceptionMamba block allows efficient computation of large-kernel depthwise convolutions via a parallel multi-branch structure inherited from the InceptionNeXt. Different from the predecessor which makes use of one-dimensional strip convolutions, our method features orthogonal band convolutions with improved local modeling capacity in characterizing multi-dimensional visual cues. Additionally, the GlobalMixer module with bottleneck Mamba structure can capture long-range dependencies through cross-channel interaction, which is conducive to boosting global modeling capability with expanded receptive field. In contrast, both ConvNeXt and InceptionNeXt demonstrate limited global modeling capacity, lacking GlobalMixer-like component for sufficient information fusion.

\section{Experiments}
To evaluate the proposed InceptionMamba architecture, we have carried our extensive experiments for classification and other downstream vision tasks including object detection and semantic segmentation. 

\subsection{Image classification on ImageNet}

\paragraph{\textbf{Experimental setup}}For classification, the public ImageNet-1k~\cite{deng2009imagenet, russakovsky2015imagenet} dataset is involved in our evaluations. Our model is trained from scratch at 224$\times$224 resolution for 300 epochs using the AdamW~\cite{loshchilov2019decoupledweightdecayregularization} optimizer with a weight decay of 5e-2, initial learning rate of 1e-3, and batch size of 512. The training strategy incorporates a cosine scheduler with 10 warmup epochs, Label Smoothing of 0.1, stochastic depth, and RandAugment. For fairness, we adopt conventional data augmentation methods including Mixup, random erasing, and AutoAugment following~\cite{mobilemamba}. Numerous advanced state-of-the-art backbone networks are involved in our comparative studies, including Vision-RWKV~\cite{duan2024vrwkv}, Vision Mamba, Vision LSTM~\cite{alkin2024visionlstm}, PlainMamba~\cite{Yang_2024_BMVC}, EfficientVMamba~\cite{pei2024efficientvmamba}, VMamba, InceptionNeXt, MambaVision, QuardMamba~\cite{xie2024quadmamba}, PeLK~\cite{chen2024pelk}, MambaOut and RDNet~\cite{kim2024densenets}. Parallel training is performed with eight NVIDIA RTX 4090 GPUs in all our experiments.

\paragraph{\textbf{Results}}As shown in Table~\ref{Tab:2}, our proposed InceptionMamba demonstrates consistent performance advantages across all model scales while maintaining superior computational efficiency. Among tiny models, InceptionMamba-T achieves 83.1\% Top-1 accuracy with only 25.4M parameters and 4.0G FLOPs, outperforming both Mamba and CNN models like VMamba-T (82.2\%) and InceptionNeXt-T (82.3\%). This trend continues in small and base variants, demonstrating that InceptionMamba-S achieves 84.1\% Top-1 accuracy on par with MambaOut-S with lower computational costs (7.8G vs 9.0G FLOPs). Analogously, InceptionMamba-B achieves state-of-the-art performance, reporting unrivaled 84.7\% accuracy with only 83M network parameters and 14.3G FLOPs. These results validate the superiority of our InceptionMamba in accuracy-efficiency trade-off compared to existing mainstream backbone networks.

\begin{table}
\caption{Performance comparison on ImageNet-1k for classification across three model scales, i.e., Tiny(T), Small(S) and Base(B).}
\resizebox{1\textwidth}{!}{
\begin{tabular}{cccccc}
\toprule
Model& Pub. & Reso. &Params(M) &FLOPs(G) &Top-1(\%) \\
\midrule
VRWKV-S~\cite{duan2024vrwkv} & ICLR 2025 & 224 & 24 & 4.6 & 80.1\\
Vim-S~\cite{vim} & ICML 2024 & 224 & 26 & 5.1 & 80.5\\
ViL-S~\cite{alkin2024visionlstm} & ICLR 2025 & 224 & 23 & 5.1 & 81.5 \\
PlainMamba-L2~\cite{Yang_2024_BMVC} & BMVC 2024 & 224 & 25 & 8.1 & 81.6 \\
EfficientVMamba-B~\cite{pei2024efficientvmamba} & AAAI 2025 & 224 & 33 & 4.0 & 81.8 \\
VMamba-T~\cite{liu2024vmamba} & NIPS 2024 & 224 & 22 & 5.6 & 82.2 \\
InceptionNeXt-T~\cite{yu2024inceptionnext} & CVPR 2024 & 224 & 28 & 4.2 & 82.3 \\
MambaVision-T~\cite{hatamizadeh2025mambavision} & CVPR 2025 & 224 & 32 & 4.4 & 82.3 \\
QuardMamba-S~\cite{xie2024quadmamba} & NIPS 2024 & 224  & 31 & 5.5 & 82.4 \\
PeLK-T~\cite{chen2024pelk} & CVPR 2024 & 224 & 29 & 5.6 & 82.6 \\
MambaVision-T2~\cite{hatamizadeh2025mambavision} & CVPR 2025 & 224 & 35 & 5.1 & 82.7 \\
MambaOut-T~\cite{yu2025mambaout} & CVPR 2025 & 224 & 27 & 4.5 & 82.7 \\
RDNet-T~\cite{kim2024densenets} & ECCV 2024 & 224 & 24 & 5.0 & 82.8 \\
\textbf{InceptionMamba-T} & Ours & 224 & 25 & 4.0 & \textbf{83.1} \\
\midrule
PlainMamba-L3~\cite{Yang_2024_BMVC} & BMVC 2024 & 224 & 50 & 14.4 & 82.3 \\
MambaVision-S~\cite{hatamizadeh2025mambavision} & CVPR 2025 & 224 & 50 & 7.5 & 83.3 \\
InceptionNeXt-S\cite{yu2024inceptionnext} & CVPR 2024 & 224 & 49 & 8.4 & 83.5 \\
VMamba-S~\cite{liu2024vmamba} & NIPS 2024 & 224 & 50 & 8.7 & 83.6 \\
RDNet-S~\cite{kim2024densenets} & ECCV 2024 & 224 & 50 & 8.7 & 83.7 \\
QuardMamba-B~\cite{xie2024quadmamba} & NIPS 2024 & 224  & 50 & 9.3 & 83.8 \\
PeLK-S~\cite{chen2024pelk} & CVPR 2024 & 224 & 50 & 10.7 & 83.9 \\
MambaOut-S~\cite{yu2025mambaout} & CVPR 2025 & 224 & 48 & 9.0 & 84.1 \\
\textbf{InceptionMamba-S} & Ours & 224 & 46 & 7.8 & \textbf{84.1} \\
\midrule
Vim-B~\cite{vim} & ICML 2024 & 224 & 98 & - & 81.9 \\
VRWKV-B~\cite{duan2024vrwkv} & ICLR 2025 & 224 & 94 & 18.2 & 82.0 \\
ViL-B~\cite{alkin2024visionlstm} & ICLR 2025 & 224 & 89 & 18.6 & 82.4 \\
VMamba-B~\cite{liu2024vmamba} & NIPS 2024 & 224 & 89 & 15.4 & 83.9 \\
InceptionNeXt-B~\cite{yu2024inceptionnext} & CVPR 2024 & 224 & 87 & 14.9 & 84.0 \\
MambaVision-B~\cite{hatamizadeh2025mambavision} & CVPR 2025 & 224 & 98 & 15.0 & 84.2 \\
PeLK-B~\cite{chen2024pelk} & CVPR 2024 & 224 & 89 & 18.3 & 84.2 \\
MambaOut-B~\cite{yu2025mambaout} & CVPR 2025 & 224 & 85 & 15.8 & 84.2 \\
RDNet-B~\cite{kim2024densenets} & ECCV 2024 & 224 & 87 & 15.4 & 84.4 \\
\textbf{InceptionMamba-B} & Ours & 224 & 83 & 14.3 & \textbf{84.7} \\
\bottomrule
\end{tabular}\label{Tab:2}
}
\end{table}

\subsection{Object detection $\&$ instance segmentation on COCO}


\paragraph{\textbf{Experimental setup}}For object detection and instance segmentation, we have conducted experiments on MS-COCO~\cite{lin2014COCO} which contains 118k training images and 5k validation images. The detection framework is utilized as Mask R-CNN~\cite{he2017mask} with our InceptionMamba backbone, and training the detector is initialized with the model pretrained on the ImageNet. In implementation, we follow the standard 1$\times$ training protocol (12 epochs), and adopt multi-scale data augmentation. To maintain the original aspect ratio, the input image is resized such that the shorter side is fixed at 800 pixels while the longer side does not exceed 1333 pixels. Using the AdamW optimizer, initial learning rate is set to 0.0002, whilst weight attenuation coefficient is 0.05. In comparative studies, competing architectures range from CNN and Transformer like ConvNeXt and Swin-Transformer to Mamba-based networks, e.g., EfficientVMamba~ and MambaOut.

\begin{table}
\centering
\caption{Performance comparison on COCO for object detection and instance segmentation. Mask R-CNN is used as detection framework.} 
\begin{tabular}{ccccc}
\toprule
Backbone & Params(M) & FLOPs(G) & $AP^{b}$ & $AP^{m}$ \\
\midrule
ConvNeXt-T~\cite{liu2022convnet} & 48 & 262 & 44.2 & 40.1 \\
Swin-T~\cite{liu2021Swin} & 48 & 267 & 42.7 & 39.3 \\
ViT-Adapter-S~\cite{chen2022vitadapter} & 48 & 403 & 44.7 & 39.9 \\
PVTv2-B2~\cite{wang2021pvtv2} & 45 & 309 & 45.3 & 41.2 \\
EfficientVMamba-B~\cite{pei2024efficientvmamba} & 53 & 252 & 43.7 & 40.2 \\
PlainMamba-L1~\cite{Yang_2024_BMVC} & 31 & 388 & 44.1 & 39.1 \\
MambaOut-T~\cite{yu2025mambaout} & 43 & 262 & 45.1 & 41.0 \\
\textbf{InceptionMamba-T} & 43 & 233 & \textbf{46.0} & \textbf{41.8} \\
\midrule
ConvNeXt-S~\cite{liu2022convnet} & 70 & 348 & 45.4 & 41.8 \\
Swin-S~\cite{liu2021Swin} & 69 & 354 & 44.8 & 40.9 \\
PVTv2-B3~\cite{wang2021pvtv2} & 65 & 397 & 47.0 & 42.5 \\
MambaOut-S~\cite{yu2025mambaout} & 65 & 354 & 47.4 & 42.7 \\
\textbf{InceptionMamba-S} & 63 & 301 & \textbf{47.5} & \textbf{42.6} \\
\midrule
ConvNeXt-B~\cite{liu2022convnet} & 108 & 486 & 47.0 & 42.7 \\
Swin-B~\cite{liu2021Swin} & 107 & 496 & 46.9 & 42.3 \\
ViT-Adapter-B~\cite{chen2022vitadapter} & 102 & 557 & 47.0 & 41.8 \\
PVTv2-B5~\cite{wang2021pvtv2} & 102 & 557 & 47.4 & 42.5\\
MambaOut-B~\cite{yu2025mambaout} & 100 & 495 & 47.4 & 43.0 \\
\textbf{InceptionMamba-B} & 99 & 421 & \textbf{48.1} & \textbf{43.1} \\
\bottomrule
\end{tabular}\label{Tab:3}
\end{table}

\paragraph{\textbf{Results}}As illustrated in Table~\ref{Tab:3}, our evaluations demonstrate superior performance of InceptionMamba across different model scales on COCO dataset. To be specific, InceptionMamba-T achieves 46.0\% $AP^b$ and 41.8\% $AP^m$ with only 43M parameters and 233G FLOPs, which consistently beats other competitors. In particular, InceptionMamba-T surpasses MambaOut-T by almost 1\% $AP^b$ with superior computational efficiency with 12.5\% less FLOPs. This advantage against ConvNeXt-T is elevated to 1.8\% $AP^b$ with 10.4\% less network parameters and 12.5\% less FLOPs. At larger model scale, our InceptionMamba-B reports the highest 48.1\% $AP^b$ and 43.1\% $AP^m$, which outperforms MambaOut-B and ConvNeXt-B while reducing computational costs by approximately 13\% and 15\%. In particular, our InceptionMamba achieves better performance than other typical Mamba-based architectures including PlainMamba, EfficientVMamba and MambaOut, sufficiently exhibiting the advantage of our framework in balancing local-aware spatial encoding and global context modeling for capturing long-range dependencies.

\begin{table}
\centering
\caption{Comparison of semantic segmentation results on ADE20K dataset using UperNet.}
\begin{tabular}{cccc}
\toprule
Backbone & Params(M) & FLOPs(G) & mIoU(\%) \\
\midrule
ConvNeXt-T~\cite{liu2022convnet} & 60 & 939 & 46.0  \\
Swin-T~\cite{liu2021Swin} & 60 & 945 & 44.4  \\
Focal-T~\cite{yang2022focal} & 62 & 998 & 45.8  \\
MambaVision-T~\cite{hatamizadeh2025mambavision} & 55 & 945 & 46.0 \\
EfficientVMamba-B~\cite{pei2024efficientvmamba} & 65 & 930 & 46.5 \\
VMamba-T~\cite{liu2024vmamba} & 55 & 964 & 47.3  \\
MambaOut-T~\cite{yu2025mambaout} & 54 & 938 & 47.4  \\
\textbf{InceptionMamba-T} & 53 & 928 & \textbf{47.3}  \\ 
\midrule
ConvNeXt-S~\cite{liu2022convnet} & 82 & 1027 & 48.7  \\
Swin-S~\cite{liu2021Swin} & 81 & 1038 & 47.6  \\
Focal-S~\cite{yang2022focal} & 85 & 1130 & 48.0  \\
MambaVision-S~\cite{hatamizadeh2025mambavision} & 84 & 1135 & 48.2 \\
VMamba-S~\cite{liu2024vmamba} & 76 & 1081 & 49.5  \\
MambaOut-S~\cite{yu2025mambaout} & 76 & 1032 & 49.5  \\
\textbf{InceptionMamba-S} & 73 & 1006 & \textbf{49.2} \\
\midrule
ConvNeXt-B~\cite{liu2022convnet} & 122 & 1170 & 49.1  \\
Swin-B~\cite{liu2021Swin} & 121 & 1188 & 48.1  \\
Focal-B~\cite{yang2022focal} & 126 & 1354 & 49.0  \\
MambaVision-B~\cite{hatamizadeh2025mambavision} & 126 & 1342 & 49.1  \\
VMamba-B~\cite{liu2024vmamba} & 110 & 1226 & 50.0 \\ 
MambaOut-B~\cite{yu2025mambaout} & 112 & 1178 & 49.6 \\
\textbf{InceptionMamba-B} & 110 & 1145 & \textbf{50.1} \\
\bottomrule
\end{tabular}\label{Tab:4}
\end{table}

\subsection{Semantic segmentation on ADE20K}
\paragraph{\textbf{Experimental setup}}In addition to the above-mentioned evaluations, we perform semantic segmentation experiments on public ADE20K dataset~\cite{zhou2017scene} using UperNet architecture~\cite{xiao2018unified, mmseg2020}. This dataset has 150 semantic categories with 20,000 training images and 2,000 validation images. Initialized with the pretrained model similar to the detection experiments, we follow \cite{yu2025mambaout} to adopt a protocol of 160,000 iterations in our training process, and utilize the AdamW optimizer with an initial learning rate set to 0.00012 and a batch size of 16.

\paragraph{\textbf{Results}}Table~\ref{Tab:4} presents the performance of various architectures on ADE20K dataset. It is shown that our tiny and small InceptionMamba models report respective 47.3\% and 49.2\% mIoU with superior efficiency, exhibiting competitive performance compared to corresponding VMamba and MambaOut competitors. When extending our framework to larger scales, the resulting InceptionMamba-B achieves 50.1\% mIoU with 110M parameters and 1145G FLOPs, which outperforms all the other competing networks in both accuracy and efficiency. These comparisons further reveal the potential of our hybrid architecture in combining local-aware encoding and global context modeling.

\subsection{Ablation studies}

\paragraph{\textbf{Branch ratio}}
To investigate the impact of different branch allocation ratios on model performance within ConvMixer, we have conducted ablation experiments comparing three different channel allocation strategies in our model excluding GlobalMixer module. The experimental results as shown in Table~\ref{Tab:5} demonstrate that the best 81.9\% Top-1 accuracy is reported when employing a ratio of [0.125, 0.125, 0.75] with 21.4M parameters and 3.35G FLOPs, exhibiting the optimal balance between accuracy and efficiency. In contrast, increasing the ratio of the second branch to 0.25 ([0.125, 0.25, 0.625]) leads to a slight parameter increase to 21.5M and a marginal performance drop (81.7\%), indicating that the asymmetric convolutional branches are sensitive to channel variations. A balanced allocation scheme ([0.25, 0.25, 0.5]) maintains 81.9\% accuracy with comparable computational cost (3.38G FLOPs), verifying the robustness of the module design to different allocation ratios. After comprehensive evaluation, the branch ratio is empirically set as [0.125, 0.125, 0.75] in our remaining experiments, as it achieves the best accuracy-efficiency trade-off.

\begin{table}[t]
\centering
\caption{Comparison Results of different settings of Branch Ratios (square conv./Band conv./Identity Mapping) within ConvMixer. The GlobalMixer is excluded from our InceptionMamba.}
\begin{tabular}{cccc}
\toprule
Branch ratio &Params(M) &FLOPs(G) &Top-1(\%) \\
\midrule
\texttt{[0.125, 0.125, 0.75]} & 21.4 & 3.35 & 81.9 \\
\texttt{[0.125, 0.25, 0.625]} & 21.5 & 3.37 & 81.7 \\
\texttt{[0.25, 0.25, 0.5]} & 21.5 & 3.38 & 81.9 \\
\bottomrule
\end{tabular}\label{Tab:5}
\end{table}

\paragraph{\textbf{ConvMixer}}
To delve into the ConvMixer within our InceptionMamba, we have conducted comprehensive ablation studies by comparing our ConvMixer with two different designs: Depthwise convolutions within ConvNeXt and decomposed convolutions within InceptionNeXt. We also replaced band convolution with strip convolution for comparison. As demonstrated in Table~\ref{Tab:6}, our proposed ConvMixer achieves superior performance while maintaining identical model complexity.  
\begin{table}[t]
\centering
\caption{Comparison Results of Different structures within our ConvMixer.}
\begin{tabular}{ccccc}
\toprule
ConvMixer &Params(M) &FLOPs(G) &Top-1(\%)  \\
\midrule
DWConv3$\times$3 & 25.4 & 4.0 & 82.8 \\ 
InceptionDWConv2d & 25.4 & 4.0 & 82.9\\
Strip convolution & 25.4 & 4.0 & 83.0 \\
\textbf{Ours} & \textbf{25.4} & \textbf{4.0} & \textbf{83.1} \\
\bottomrule
\end{tabular}\label{Tab:6}
\end{table}

\paragraph{\textbf{Necessity of Bottleneck}}
Moreover, we explore the benefits of bottleneck structure within our GlobalMixer. As shown in Table~\ref{Tab:7}, integrating the bottleneck with SS2D maintains model accuracy while significantly improving efficiency. Compared to the baseline SS2D, introducing the bottleneck achieves the same Top-1 accuracy while reducing both parameters and computational costs by over 10\%. Additionally, the throughput increases substantially, highlighting the ability of the bottleneck in efficiently compressing feature dimensions without sacrificing model performance as aforementioned.
\begin{table}[t]
\centering
\caption{Comparison of SS2D (Mamba) and Bottleneck + SS2D (GlobalMixer).}
\begin{tabular}{ccccc}
\toprule
Global mixer & Params(M) & FLOPs(G) & Top-1(\%) & TP.(imgs/s) \\
\midrule
SS2D & 29 & 4.6 & 83.1 & 1042 \\ 
\textbf{Bottleneck + SS2D} & \textbf{26} & \textbf{4.0} & \textbf{83.1} & \textbf{1362} \\
\bottomrule
\end{tabular}\label{Tab:7}
\end{table}

\paragraph{\textbf{GlobalMixer}}
To further investigate the GlobalMixer within our framework, we have systematically evaluated different architectures involving the bottleneck as illustrated in Table~\ref{Tab:8}. Compared to the baseline without Mamba Bottleneck, all the other variants involving bottleneck improve Top-1 accuracy. In particular, our GlobalMixer achieves the highest 83.1\% Top-1 accuracy, while maintaining competitive efficiency by reducing FLOPs by 13\% compared to the structure of Bottleneck $+$ attention (LiteMLA linear attention in EfficientViT~\cite{cai2023efficientvit}) and promoting the throughput from 1305 imgs/s to 1362 imgs/s. Notably, while Bottleneck $+$ attention slightly improves accuracy, it leads to significantly increased computational cost. This contrast suggests the advantage of SS2D in balancing between performance and efficiency. Simpler designs like Bottleneck $+$ GELU or Bottleneck $+$ DWConv7×7 achieve marginal performance gains but fail to rival SS2D in terms of global modeling capability. These results demonstrate that our GlobalMixer incorporating SS2D and bottleneck optimally leverages hardware-friendly operations (e.g., linear complexity scans) to achieve accuracy-speed trade-off, thereby well aligning with the efficiency goal of modern lightweight architectures.
\begin{table}[t]
\centering
\caption{Comparison of different modules involving Bottleneck structure. }
\scalebox{0.9}{
\begin{tabular}{ccccc}
\toprule
Global mixer & Params(M) & FLOPs(G) & Top-1(\%) & TP.(imgs/s) \\
\midrule
no Bottleneck Mamba & 21.4  & 3.3  &  81.9 & 2255 \\
Bottleneck + GELU & 23.5 & 3.6 & 82.1 & 1976\\
Bottleneck + DWConv7$\times$7 & 23.6 & 4.0 & 82.1 & 1880 \\ 
Bottleneck + attention & 26.4 & 4.6 & 82.3  & 1305\\
\textbf{Bottleneck + SS2D} & \textbf{25.4} & \textbf{4.0} & \textbf{83.1} & \textbf{1362}\\
\bottomrule
\end{tabular}}\label{Tab:8}
\end{table}

\begin{figure}
    \centering
       \includegraphics[width=1.0\textwidth]{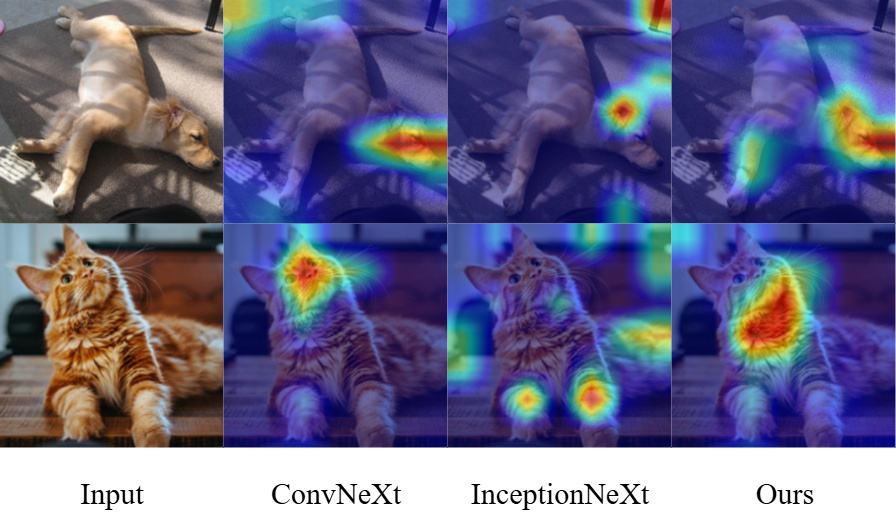}  
    \caption{Comparison of CAMs generated from different mainstream architectures. Our InceptionMamba is capable of characterizing key semantic-aware regions, which is beneficial for a variety of downstream vision tasks.}
    \label{Fig:6}
\end{figure}

\paragraph{\textbf{More Visualization Results}}
To qualitatively demonstrate the advantages of the proposed InceptionMamba, we intuitively compare our method with the classic ConvNeXt and InceptionNeXt frameworks. As illustrated in Fig.~\ref{Fig:6}, given an input ``dog'' image, InceptionMamba produces CAMs that can capture object-aware regions more comprehensively and accurately, leading to high responses to the regions of the dog's limbs and head. In contrast, for the input ``cat'' image, ConvNeXt mainly focuses on the face of the cat, while InceptionNeXt pays attention to the cat's legs area. In contrast, our InceptionMamba can capture key regions including the cat's head and body area that are highly semantically relevant to the ``cat'' class. This indicates that our InceptionMamba architecture can improve global context modeling capacity while inherits multi-scale feature fusion from Inception-like structures. Besides, the qualitative comparisons demonstrate that InceptionMamba can capture richer spatial relationships, manifesting itself as a potential state-of-the-art lightweight architecture. This also complements previous ablation results regarding the advantage of our method in efficiency-accuracy trade-off.


\section{Conclusion}

In this study, we propose a hybrid backbone architecture termed InceptionMamba. Inheriting efficient parallel structure from InceptionNeXt framework, we leverage large band depthwise convolutions within our ConvMixer for capturing local cues and develop bottleneck Mamba for channel interaction, such that long-range dependencies modeling capability can be significantly improved. Extensive experiments demonstrate the superiority of InceptionMamba in terms of both accuracy and computational efficiency compared to existing mainstream CNNs and ViT architectures on various public benchmarks spanning image classification and several downstream vision tasks. We hope our work can inspire the exploration of efficient multi-scale global interaction paradigms in deep architecture design and facilitate the development of lightweight yet powerful vision backbones.

\section*{Acknowledgment}
This work was supported by the National Natural Science Foundation of China under Grant 62173186, 62076134, 62276061, and the Key R \& D Program of Zhejiang Province (2024C04056(CSJ)).

\bibliographystyle{elsarticle-num} 
\bibliography{ref}







\end{sloppypar}

\end{document}